\newcommand{\longversion}[1]{#1}
\newcommand{\shortversion}[1]{}

\documentclass[letterpaper]{article}
\usepackage{aaai23} 
\usepackage{times} 
\usepackage{helvet} 
\usepackage{courier} 
\usepackage[hyphens]{url} 
\usepackage{graphicx} 
\usepackage{graphicx} 
\usepackage{natbib} 
\usepackage{caption} 
\usepackage{siunitx}
\usepackage{multirow}
\usepackage{mathtools}

\usepackage{multicol}

\usepackage{amsmath,amsfonts,amsthm,bm}
\usepackage{amssymb}

\usepackage{pifont}
\newcommand{\cmark}{\ding{51}}%
\newcommand{\xmark}{\ding{55}}%
\usepackage{lipsum,adjustbox}
\usepackage{url}
\usepackage{mathtools}
\usepackage{xfrac}
\usepackage{makecell}
\usepackage{rotating}
\usepackage[tableposition=below,belowskip=-5pt,aboveskip=5pt]{caption}
\usepackage{balance}
\usepackage{amsthm}

\usepackage{dot2texi}
\usepackage{amsmath}
\usepackage{amsthm}
\usepackage{thmtools, thm-restate}
\usepackage{amsfonts}
\usepackage{tikz}
\usetikzlibrary{shapes,arrows,automata}
\usepackage{listings}

\usepackage{fancybox}
\usepackage{enumerate}
\usepackage{booktabs}

\usepackage{subfig}
\newtheorem{theorem}{Theorem}
\definecolor{cyan}{rgb}{0,1,1}

\definecolor{black}{rgb}{0,0,0}
\newcommand{\black}[1]{\textcolor{black}{#1}}

\newtheorem{definition}{Definition}

\newtheorem{problem}{Problem}

\title{STL-Based Synthesis of Feedback Controllers Using Reinforcement Learning}

\author{
    Nikhil Kumar Singh and Indranil Saha \\
}

\affiliations {
    Department of Computer Science and Engineering\\
    IIT Kanpur\\
    \{nksingh, isaha\}@cse.iitk.ac.in
}

\begin{document}

\maketitle

\begin{abstract}
Deep Reinforcement Learning (DRL) has the potential to be used for synthesizing feedback controllers (agents) for various complex systems with unknown dynamics.
These systems are expected to satisfy diverse safety and liveness properties best captured using temporal logic.
In RL, the reward function plays a crucial role in specifying the desired behaviour of these agents. 
However, the problem of designing the reward function for an RL agent to satisfy complex temporal logic specifications has received limited attention in the literature.
To address this, we provide a systematic way of generating rewards in real-time by using the quantitative semantics of Signal Temporal Logic (STL), a widely used temporal logic to specify the behaviour of cyber-physical systems. 
We propose a new quantitative semantics for STL having several desirable properties, making it suitable for reward generation. 
We evaluate our STL-based reinforcement learning mechanism on several complex continuous control benchmarks and compare our STL semantics with those available in the literature in terms of their efficacy in synthesizing the controller agent. 
Experimental results establish our new semantics to be the most suitable for synthesizing feedback controllers for complex continuous dynamical systems through reinforcement learning.
\end{abstract}


\section{Introduction}

Feedback controllers form the core of any safety-critical cyber-physical systems (CPSs). 
The traditional approaches for synthesizing feedback controllers rely on the availability of a mathematical model of the dynamical system whose behaviour the controller is supposed to regulate.
However, for complex dynamical systems, the creation of a faithful mathematical model poses a tremendous challenge to the control engineers.
Reinforcement learning (RL) provides an alternative for synthesizing feedback controllers in the form of a controller agent for complex systems without precise mathematical models. 
The agent interacts with the dynamical system in a simulation environment providing a faithful but complex system model that is not suitable for controller synthesis using traditional control-theoretic procedures.

A deep neural network can be used as the agent in RL, and in that case, the learning procedure is called the \emph{Deep Reinforcement Learning} (DRL).
In recent times, Deep Reinforcement Learning (DRL) has been extremely popular in solving highly complex problems such as solving Atari~\cite{Mnih15} and Go~\cite{Silver17}, making BiPedal Robots walk~\cite{lillicrap19}, learning visuomotor controllers for robots~\cite{Levine16}, and many more. The success of DRL in solving those complex problems is attributed to well-defined reward functions. 
Thus, designing correct reward functions~\cite{sutton} is extremely important for synthesizing DRL-based controllers. 

Several papers have considered the RL-based methods for synthesizing feedback controllers (e.g.~\cite{lillicrap19,levine13,Deisenroth13,Fulton18,Hafner2011}).
In this approach, the reward function is designed by a control engineer having complete knowledge of the system dynamics. 
As the system becomes high-dimensional and highly nonlinear, designing a correct reward function in this manner becomes prohibitively hard. 
For control engineers, it would have been significantly more convenient if they could write the specification of the closed-loop system in a formal language, and the reward could be generated automatically from this specification.

In the recent past, Signal Temporal Logic (STL)~\cite{donze10a} has been used widely in capturing real-time specifications for synthesizing controllers for complex CPSs~\cite{Singh21,raman14,raman15}.
The robustness semantics of STL makes it a potential candidate for being used for specification-based reward generation in  controller synthesis through DRL.
An STL-based reward can efficiently perform \emph{temporal aggregation}, which is difficult to achieve in a hand-crafted reward function.
However, the classical quantitative semantics~\cite{donze10a},~\cite{jaksic18} of STL often leads to improper rewards, which in turn may lead to the synthesis of sub-optimal controllers. This is because the point-based classical semantics suffer from the \emph{shadowing problem}~\cite{varnai20}, 
where an increase in the robustness of an  individual sub-specification does not influence the robustness of the full specification computed by the AND operator, except for the case when it is minimum.

Of late, researchers have proposed several new semantics of STL that attempt to incorporate aggregate notions of robustness instead of point-based estimates provided by the classical semantics. 
Some of these semantics such as \texttt{AGM}~\cite{agm} and \texttt{SoftMax}~\cite{varnai20} were designed to address the \emph{shadowing problem}.
However, as the robustness functions for these semantics are not smooth, they are not well-suited for RL-based controller synthesis.
Another semantics \texttt{LSE}~\cite{lse,Pant17} aims to provide a smooth approximation to the robustness function, but it does not address the shadowing problem.
Due to the limitations of the existing aggregation-based semantics, STL has not yet been adopted for DRL-based controller synthesis despite its tremendous potential. 

In this work, we propose a new aggregate-based semantics for STL, called \texttt{SSS} (Smoothened-Summation-Subtraction).
Our Semantics is not sound, but we show that soundness is not an essential requirement of an STL semantics when it is used for reward generation in the DRL process. 
Rather, our semantics have all the essential properties for being qualified as a reward generation mechanism --- it offers smoothness, addresses the shadowing problem, and ensures min-max boundedness. 

We implement our semantics in the online monitoring tool RTAMT~\cite{rtamt} and evaluate it on several challenging continuous control benchmarks available in the \texttt{gym} environment~\cite{gym}. For each of those benchmarks, we introduce a suitable STL specification that captures the safety and liveness requirements of the system. 
We compare the performance of the controller synthesized using our semantics with those synthesized with other aggregate-based semantics available in the literature. Experimental results establish that our semantics consistently outperforms all the other semantics by a significant margin.

\section{Preliminaries}

\subsection{Closed-loop Control System}

Open-loop dynamical systems (a.k.a. plants) often do not satisfy desired specifications.
Control engineers design a feedback controller $\mathcal{C}$ for an open-loop system  $\mathcal{P}$ to ensure that the 
closed-loop system $\mathcal{M}$ satisfies its specification.
We assume that the set of states of the closed-loop system is fully observable and thus is the same as the set of outputs, denoted by $\mathcal{X}$.
The output of the system ($x_t$) is used to generate control input ($u_t$) that is applied to the system at every time step $t$ to regulate its behavior.
A trace $\omega = \langle x_0,u_0\rangle,\langle x_1,u_1\rangle,\ldots$ is defined as the sequence of alternating states and control inputs of the system evolving in discrete time-steps,
where $x_t$ and $u_t$, \mbox{$i \in \mathbb{N}$}, refer to the state (output) and control input at the $t$-th time-step.
We use $\mathcal{L}(\mathcal{M})$ to denote the set of all traces of $\mathcal{M}$.

\subsection{Signal Temporal Logic}
Signal Temporal Logic (STL)~\cite{donze10a} 
enables us to reason about the real-time properties of signals (simulation traces). These specifications consist of real-time predicates over the signal variables.

The syntax of an STL specification $\phi$ is defined by the grammar
\begin{equation}
   \phi = \mathsf{true} \mid \pi \mid \neg \phi \mid \phi_{1} \land \phi_{2} \mid \phi_{1} \,U_I\, \phi_{2}, 
\end{equation}
where $\pi \in \Pi$, $\Pi$ is a set of atomic predicates defined on the outputs of the closed-loop system, and $I$ $\subseteq$ $\mathbb{R^+}$ is an arbitrary interval of non-negative real numbers. The  operators $\neg$ and $\land$ denote logical NOT and AND operators. Other logical operators like logical OR ($\vee$) and implication ($\implies$) can be derived using $\neg$ and $\land$. The temporal operator $U_I$ is the \emph{until} operator implying that $\phi_2$ becomes $\mathsf{true}$ sometime in the time interval $I$ and $\phi_1$ must remain $\mathsf{true}$ until $\phi_2$ becomes $\mathsf{true}$.
There are two other useful temporal operators, namely  \emph{eventually} ($\lozenge_I$) and \emph{always} ($\square_I$), which can be derived from the temporal and logical operators defined above. The formula $\lozenge_I\, \phi$ means that the formula $\phi$ will be $\mathsf{true}$ sometime in the time interval $I$. The formula $\square_I\, \phi$ means that the formula $\phi$ will always be $\mathsf{true}$ in the time interval $I$.

\smallskip
\noindent
\textit{Robust Semantics of STL}: 
We follow the robust semantics of STL provided in \cite{Donxe13}. We use Euclidean metric as the norm to measure the distance $d$ between two values $v, v' \in \mathbb{R}$, i.e., $d(v,v') = \parallel v-v' \parallel$. Let $v \in \mathbb{R}$ be a value, $A \subseteq \mathbb{R}$ be a set. Then the \emph{signed distance} from $v$ to $A$ is defined as:
\begin{equation}
    \mathtt{Dist }(v,A)= 
  \begin{cases}
      \ \ \ inf\{d(v,v')\ |\ v' \notin A\} & \text{if } v \in A, \\
      -inf\{d(v,v')\ |\ v' \in A\} & \text{if } v \notin A.
   \end{cases}
\end{equation}

Intuitively, $\mathtt{Dist }(v,A)$ captures how far a value $v$ is from the violation of the inclusion in the set $A$.
In both cases, we search for the minimum distance between $v$ and a point on the boundary of $A$. 
Also, the case $v \notin A$ refers to a violation. Thus, the negative sign is used in the definition.

We use $\mathsf{O}:\mathcal{K} \rightarrow 2^{\mathcal{X}}$ to denote the mapping of a predicate $\kappa$ to a set of states ($\mathcal{X}$).
Given a trace $\omega$ and the mapping $\mathsf{O}$, we define the robust semantics of $\omega$ w.r.t. $\phi$ at time $t \in \mathbb{R}$, 
denoted by $\rho(\phi,\omega,t)$, by induction as follows:
\begin{small}
\begin{subequations}
\label{stlrobeq}
\begin{align}
    &\rho(\mathsf{true},\omega,t) \ \ \ \ \ =  \ +\infty, \\
    &\rho(\kappa,\omega,t) \ \ \ \ \ \ \ \ \ = \ \mathtt{Dist }(x_t,\mathsf{O}(\kappa)), \\
    &\rho(\neg \phi,\omega,t) \ \ \ \ \ \ =  \ - \rho(\phi,\omega,t), \\
    &\rho(\phi \land \psi,\omega,t) \ \, =  \ \mathtt{min}(\rho(\phi,\omega,t),\rho(\psi,\omega,t)), \label{stland}\\
    &\rho(\phi \, U_{I} \, \psi,\omega,t) =  \ \sup_{t' \in t + I} \ \mathtt{min} (\rho(\psi,\omega,t'), \inf_{t'' \in [t, t']} \rho(\phi,\omega,t'')).
\end{align}
\end{subequations}
\end{small}

The robustness of a trace $\omega$ w.r.t a specification $\phi$ is defined as $\rho(\phi,\omega)=\rho(\phi,\omega,0)$, i.e., the robustness at time $0$. 
If $\rho(\phi,\omega,t)\ne 0$, its sign indicates the satisfaction status. 
The robustness metric $\rho$ maps each simulation trace $\omega$ to a real number $\rho$.
 Intuitively, the robustness of a trace $\omega \in \mathcal{L}(\mathcal{M})$ with respect to an STL formula $\phi$ is the radius of the largest ball centered at trace $\omega$ that we can fit within $\mathcal{L_\phi}$, where $\mathcal{L_\phi}$ is the set of all signals satisfying $\phi$.

The semantics described above is often referred to as \textit{classical} semantics. Apart from this, other popular semantics of STL are based on Arithmetic-Geometric mean~\cite{agm}, Logarithm-summation-exponential~\cite{lse,Pant17} and Softmax~\cite{varnai20}.

\subsection{Online Monitoring of STL}
The standard computation of STL robustness is offline, i.e., the signal values over the entire time horizon are available. 
An online monitor takes as input the signal value at each time instant, and it computes the robustness of the signal at that time instant w.r.t a given specification. 
Here, the specifications can contain future temporal operators (like eventually, always, etc.), which are impossible to evaluate until the end of the episode. Hence, these specifications are converted into some equi-satisfiable form which postpones the evaluation of the formula from the current time to the end of horizon~\cite{rtamt}. Intuitively, the online monitor continuously computes the robustness of specification w.r.t. each new data coming in. 






For online robustness $\bar{\rho}$ computation, we change the definition of $U_I$ operator in the classical robustness semantics of STL (Equation~\ref{stlrobeq}).
\begin{small}
    $$\bar{\rho}(\phi \, U_{I} \, \psi,\omega,t) =  \ \sup_{t' \in t - I} \ \mathtt{min} (\bar{\rho}(\psi,\omega,t'), \inf_{t'' \in [t', t]} \bar{\rho}(\phi,\omega,t'')).$$
\end{small}
This involves the computation of robustness backward in time.
Note that for the online robustness computation, it is required that $length(\omega)\geq horizon(\phi)$.
The horizon $h$ for a specification $\phi$ is defined as follows:
\begin{subequations}
\label{stlhzneq}
\begin{align}
    h(\mathsf{true},\omega,t) \ \ \ \ & =  \ 0, \\
    h(\kappa,\omega,t) \ \ \ \ \ \ \ \ \, & = \ 0, \\
    h(\neg \phi,\omega,t)  \ \ \ \ \ \ & =  \ h(\phi,\omega,t), \\
    h(\phi \land \psi,\omega,t) \ \, & =  \ \mathtt{max}(h(\phi,\omega,t),h(\psi,\omega,t)), \\
     h(\phi \lor \psi,\omega,t) \ \, & =  \ \mathtt{max}(h(\phi,\omega,t),h(\psi,\omega,t)), \\
     h(\lozenge_I \phi,\omega,t) \ \ \ \ & =  \ I+ h(\phi,\omega,t), \\
     h(\square_I \phi,\omega,t) \ \ \ \  & =  \ I+ h(\phi,\omega,t), \\
    h(\phi \, U_{I} \, \psi,\omega,t) & =  \ I + \mathtt{max}(h(\phi,\omega,t),h(\psi,\omega,t)).
\end{align}
\end{subequations}

In this paper, the symbol $\bar{\rho}$ is used to denote robustness function only for online setting whereas $\rho$ to define the general robustness function, i.e., applicable to both online and offline setting.

\section{Problem}

This section presents the controller synthesis problem that we address in this paper. 
We also introduce a few criteria to evaluate the synthesized controllers.

\subsection{Controller Synthesis Problem}
\label{sec-prob-def}

The controller synthesis problem addressed in this paper is formally presented below.
\begin{problem}[Controller Synthesis] 
\label{prob-def}
Let $\mathcal{P}$ be an open-loop dynamical system with unknown dynamics. Let the STL specification for the system be given by $\phi$. Assuming that $\mathcal{P}$ does not satisfy $\phi$, synthesize a controller $\mathcal{C}^*$ so that the expectation of robustness of all the traces generated by the closed-loop system $\mathcal{M}$ with respect to the specification $\phi$ gets maximized.
Mathematically,
    \begin{equation}
    \mathcal{C}^*= \underset{\mathcal{C}}{\arg\max} \  \underset{\omega \sim \mathcal{C}}{\mathbb{E}} \ [\rho(\phi,\omega)].
\end{equation}
\end{problem}

\subsection{Controller Specification}


It is well established that the specification of a real system should be captured as a conjunction of safety (something bad never happens) and liveness (something good happens eventually) properties~\cite{Alpern87,Kindler94}.
 Since real systems do not run forever, the liveness requirement is specified with a time-bound and is called bounded liveness~\cite{Lamport00}.

In this paper, we consider the systems involving locomotion where the liveness specification requires the system to make steady progress towards the goal, and the safety specification requires that the system does not move to a bad state during its movement. 
For example, in the case of a walker~\cite{ant}, the liveness condition would mean that the walker always moves forward, and the safety condition would ensure that it never falls down during the movement. 
During the controller synthesis phase, we perform the synthesis with a restricted notion of liveness, i.e., we want to ensure that the system makes a certain amount of progress within a given duration. 
Our goal is to learn the best possible controller in terms of liveness while ensuring the safety constraints. 



\subsection{Controller Evaluation}
\label{sec-evaluation-metrics}

Let $T$ denote the duration of an episode for observing the behavior of the closed-loop system.
Let the number of outputs of the system be $p$ and the number of control inputs be $q$. The state output and the control input at time step $t$ are denoted by $x_t = (x_t^1, x_t^2, \ldots, x_t^p)$ and $u_t = (u_t^1, u_t^2, \ldots, u_t^q)$.
In this paper, we consider continuous control based locomotion benchmarks that have ``distance covered" as a state.
To evaluate a controller and compare it with other controllers, we use the metrics defined below.


\begin{definition}[Control Cost]
We define the control cost at a given time-step $t$ as
\begin{align*}
CC_t = \sqrt{\frac{1}{q} \cdot \sum_{i=1}^{q} (u_t^i)^2} 
\end{align*}
Now, the overall control cost (CC) is given by
\begin{equation}
CC = \frac{1}{T} \cdot \sum_{t=1}^{T} CC_t^2 
\end{equation}
\end{definition}

\begin{definition}[Distance Covered (DC)]
We define the distance covered (DC) for a full episode as
\begin{equation}
DC = x^k_T 
\end{equation}
where $x^k$ is the output corresponding to the distance.
\end{definition}

\begin{definition}[Margin of Satisfaction (MoS)]
We define the margin of satisfaction (MoS) for a  trace $\omega$ of length $T$ as
\begin{equation}
MoS = \frac{1}{T} \cdot \sum_{t=1}^{T} \rho(\phi,\omega,t) 
\end{equation}
\end{definition}

\begin{definition}[Safety Satisfaction (SAT)]
Let $\phi_s$ denote the safety component of the specification $\phi$. We define the safety satisfaction (SAT) for a full episode of length $T$ as
\begin{equation}
SAT = \mathbb{I}{[min(\{\rho(\phi_s,\omega,t)\}_{t=1}^{T})]} 
\end{equation}
where $\mathbb{I}$ is an indicator function which is $1$ when the inner expression is positive and $0$ otherwise. 
\end{definition}

Unlike these metrics, the \texttt{gym} provides a reward function for each environment which is applicable to that specific environment only. We use this metric also in our evaluaation and refer it as \textit{Default Reward (DR)}.

Ideally, a controller with low CC, high DC, high MoS, high SAT, and high DR is preferable.
Note that both MoS and SAT are evaluated using the classical semantics of STL (by convention). Since the classical semantics uses the $\min$ function for conjunction (AND), the margin of satisfaction (MoS) of the whole specification often depends on the margin of safety specification (a.k.a. safety margin). This is because even though the liveness predicate values increase, generally the safety predicate values are bounded within a range, and hence the minimum for the whole specification most of the time depends on safety specification.  






\section{Methodology}
\label{methodology}



In this section, we present the deep reinforcement learning-based controller synthesis mechanism using STL specifications. 
Our controller synthesis methodology uses STL specifications to guide the policy search. 
However, unlike the traditional guided policy search method~\cite{levine13}, we do not employ any trajectory optimization method~\cite{tassa12}. Rather, the STL specification is considered as the collection of trajectories representing the expected behaviour of the system.
The other benefit of using an STL specification is that we can utilize the quantitative semantics to generate the reward online using the robustness function.

To solve Problem~\ref{prob-def} using reinforcement learning, we represent the model-free environment  $M=\langle X,U,\mathcal{T},\bar{\rho}\rangle$,
where $X$ denotes the set of continuous states $x\in\mathbb{R}^p$ of the system, $U$ denotes the set of continuous control actions $u\in\mathbb{R}^q$ and $\bar{\rho}(\phi,\omega)\in \mathbb{R}$ represents the robustness of trace $\omega$ w.r.t. specification $\phi$. The function $\mathcal{T}:\mathbb{R}^{p\times q\times p}\rightarrow[0,\infty)$ represents the unknown dynamics of the system, i.e., the probability density of transitioning to the next state given the current state and action. 
To find the optimal policy $\pi^*: S\rightarrow A$, we define a parameterized policy $\pi_{\theta}(u|x)$ as the probability of choosing an action $u$ given state $x$ corresponding to parameter $\theta$, i.e.
\begin{equation}
\pi_{\theta}(u|x) = \mathbb{P}[u|x;\theta].
\end{equation}

We define the cost function (reward) associated with the policy parameter $\theta$ as follows:
\begin{equation}
\label{eqpd}
J(\theta) =  \underset{\omega \sim \pi_{\theta}}{\mathbb{E}} \ [\bar{\rho}(\phi,\omega)],
\end{equation}
where $\underset{\omega \sim \pi_{\theta}}{\mathbb{E}}$ denotes the expectation operation over all traces $\omega$ generated by using the policy $\pi_{\theta}$.

Our goal is to learn a policy that maximizes the reward. 
Mathematically,
\begin{equation}
\theta^* = \underset{\theta}{\arg\max} \ 
    \  J(\theta)
\end{equation}

Hence, our optimal policy would be the one corresponding to $\theta^*$, i.e., $\pi_{\theta^*}$.

The direct way to find the optimal policy is via policy gradient (PG) based algorithms. However, the PG algorithms are not sample-efficient and also unstable for continuous control problems. Actor-critic algorithms~\cite{Konda03}, on the other hand, are highly sample-efficient as they use Q-function approximation (critic) instead of sample reward return. 

The gradient for the Actor (policy) can be approximated as follows: 
\begin{equation}
\label{grad}
    \nabla_{\theta}J(\theta) = 
   {\mathbb{E}_{\pi_{\theta}}}  \
    \ [ \sum_{t=0}^{T-1}
    \ \nabla_{\theta} \ log \ \pi_{\theta}(u_t|x_t) \cdot Q(x_t,u_t)] 
\end{equation}
The function $Q(x_t,u_t)$ is given as:
\begin{equation}
\label{qfcn}
Q(x_t,u_t)= \mathbb{E}_{\pi_{\theta}}    [\sum_{t=t'}^{T} \bar{\rho}(\phi,\omega_{[t'-h(\phi),t']})|x_t,u_t]
\end{equation}
Here, $\omega_{[t-h(\phi),t]}$ denotes the partial trace over which online robustness is computed. The length of the trace is given by $h(\phi)$.


The policy gradient, given by
Equation~\eqref{grad}, is used to find the optimal policy. 
The critic, on the other hand, uses an action-value function $Q^{w}(x_t,u_t)$ to estimate the function $Q(x_t,u_t)$.
The term $\bar{\rho}(\phi,\omega_{[t'-h(\phi),t']})$ in Equation~\eqref{grad} refers to the online STL robustness, i.e., the robustness at time $t$. 

The Q value plays an important role in RL. It has a considerable influence on the types of policies that can be expressed and the region of exploration.
In fact, smoother Q-values (and hence $\rho$) are extremely useful in RL~\cite{nachum18} as it leads to smooth policies~\cite{shen20}. Moreover, using the smooth robustness function as the reward has similar semantic guarantees as the classical robustness function and provides significant speedup in learning~\cite{lse}.

\section{A New Aggregation-Based STL Semantics}

In this section, we present a new aggregation-based STL semantics and prove its various desirable properties.

In Table~\ref{table-prop}, we provide a comparison of the different  state-of-the-art STL semantics - \texttt{Classical}, \texttt{AGM}, \texttt{LSE}, and \texttt{Softmax}
along with our semantics (\texttt{SSS}) with respect to the four properties - Soundness,  Min-max boundedness, Shadow-lifting, and Smoothness. 
The classical semantics suffers from the shadowing problem and is not smooth due to the usage of $\max$/$\min$ functions. 
The \texttt{AGM} and \texttt{SoftMax} semantics address the shadowing problem but are not smooth. The \texttt{LSE} semantics, although smooth, does not address the shadowing problem. Our \texttt{SSS} semantics (introduced in the next subsection), though not sound, addresses the shadowing problem and is smooth as well.

\begin{table}[t]
\begin{center}
\medskip
\begin{tabular} {lllll}
\toprule
   STL Semantics & $SD$ & $MB$ & $SL$ & $SM$ \\
  \midrule
  $Classical$ & \cmark & \cmark & \xmark & \xmark\\
  \texttt{AGM} & \cmark & \xmark & \cmark & \xmark \\
  \texttt{LSE} & \xmark & \cmark & \xmark & \cmark\\
  \texttt{SoftMax} & \cmark & \cmark & \cmark & \xmark \\
  \texttt{SSS} & \xmark & \cmark & \cmark & \cmark\\
 \bottomrule
\end{tabular}
\caption{Comparing properties of different Semantics (SD=Soundness;  MB=MinMax-boundedness; SL=Shadow-Lifting; SM=Smoothness).}
\label{table-prop}
\end{center}
\end{table}

\subsection{Smoothened-Summation-Subtraction}

In this section, we present our new semantics \textbf{Smoothened-Summation-Subtraction (\texttt{SSS})} that has been designed specifically to be used for reward generation in the RL-based controller synthesis algorithm. Since the semantics of AND operator is sufficient to generate the full semantics, we will define the semantics only for the AND operator.




The default definition of the minimum of two numbers is given by 
\begin{equation}
\label{mindef}
 \min\,(x_1, x_2) = \frac{(x_1+x_2)-|\,x_1-x_2\,|}{2}   
\end{equation}

We extend this definition to approximate the robustness for AND as
\begin{equation}
\label{and1}
 \mathcal{A}(\rho_1,\ldots,\rho_n) = \frac{\sum_{i=1}^{n} \rho_i -|\,\underset{i}{\max\,}(\rho_i)-\underset{i}{\min\,}(\rho_i)\,|}{n} 
\end{equation}

We formulate the approximation for AND in  expression~\eqref{and1} because of two reasons - firstly, we want to use an aggregate measure of all the robustness in the AND expression, and secondly, we want to penalize the largest difference between any two robustness. Let us denote this  
largest difference as $\delta^{max}$.

In Equation~\ref{and1}, the non-smoothness of the AND operator is because of two reasons - the modulus function ($|.|$) and the max/min function.
The function $|\,x\,|$ is not smooth (differentiable) at 0 and hence can be approximated by a smooth function~\cite{Bagul20}. We found the most suitable for approximation of $|\,x\,|$ as $x.\mathtt{erf}(\mu.x)$ using the Gaussian error function $\mathtt{erf}$ where $\mu$ is a smoothness parameter. The $\mathtt{erf}$ function is defined as  
\begin{equation}
 \mathtt{erf}(x) = \frac{2}{\pi} \int_{0}^{x} e^{-t^2} dt.
\end{equation}


Hence, our approximation to the robustness for the AND expression is
\begin{small}
\begin{equation}
\label{add-sub-approx}
 \mathcal{A}(\rho_1,\ldots,\rho_n) =  \frac{\sum_{i=1}^{n} \rho_i-\{\delta^{\max}\cdot\mathtt{erf}(\mu\cdot\delta^{\max})\}}{n} 
\end{equation}
\end{small}
where $\delta^{max}=\underset{i}{\max}(\rho_i)-\underset{i}{\min\,}(\rho_i)$.

The non-smoothness w.r.t the max/min function in the $\delta^{max}$ expression is addressed as follows:
\begin{small}
\begin{equation}
\label{add-sub-approx-approx}
 \delta^{max} \approx \frac{\log(n+ \sum_{i=1}^{n}\sum_{j=1,j\ne i}^{n}\exp(\eta\cdot(\rho_i-\rho_j)))}{\eta}.
\end{equation}
\end{small}

\shortversion{
Please see the full version~\cite{fullpaper} for derivation.
}
\longversion{

As we know the LSE~\cite{lse} approximation of min and max function is given as
\begin{align*}
    \underset{i}{\min}(\rho_i) \approx -\frac{1}{\eta} \log \sum_{i=1}^{n}exp(-\eta\cdot\rho_i)
\end{align*}
\begin{align*}
    \underset{i}{\max}(\rho_i) \approx \frac{1}{\eta} \log \sum_{i=1}^{n}exp(\eta\cdot \rho_i)
\end{align*}
Hence, we can approximate $\delta^{max}=\underset{i}{\max}(\rho_i)-\underset{i}{\min\,}(\rho_i)$ with $\hat{\delta}^{max}$ given as
\begin{align*}
    \hat{\delta}^{max}
    &\approx \frac{\log(\sum_{i=1}^{n}\exp(\eta \cdot\rho_i))+ \log(\sum_{i=1}^{n}\exp(-\eta\cdot \rho_i))}{\eta}\\
    &\approx \frac{\log(\sum_{i=1}^{n}\exp(\eta\cdot\rho_i)\cdot\sum_{i=1}^{n}\exp(-\eta\cdot \rho_i))}{\eta}\\
    &\approx \frac{\log((e^{\eta\cdot\rho_1}+e^{\eta\cdot\rho_2}+...)\cdot(e^{-\eta\cdot\rho_1}+e^{-\eta\cdot\rho_2}+..))}{\eta}\\
    &\approx \frac{\log(e^{\eta\cdot(\rho_1-\rho_1)}+e^{\eta\cdot(\rho_2-\rho_2)}+...+e^{\eta\cdot(\rho1-\rho_2)}+...)}{\eta}\\
    &\approx \frac{\log(1+1+..+\sum_{i=1}^{n}\sum_{j=1,j\ne i}^{n}\exp(\eta\cdot(\rho_i-\rho_j)))}{\eta}\\
    &\approx \frac{\log(n+ \sum_{i=1}^{n}\sum_{j=1,j\ne i}^{n}\exp(\eta\cdot(\rho_i-\rho_j)))}{\eta}
  \end{align*}
}


In the above expression, $\delta^{max}$ gets a smooth approximation of the largest difference between any two robustness values $\rho_i-\rho_j$.
In the above expressions, $\mu$ and $\eta$ are tunable parameters.
Note that as $\eta \rightarrow \infty, \mu \rightarrow \infty$ reduces the AND operator in~\eqref{add-sub-approx} to the traditional $\min$ function. We define our semantics with a high $\eta$ and low $\mu$ value. A high value of $\eta$ enables us to be close to the true value of $\delta^{max}$ and a low $\mu$ gives us smoothened overall robustness value. 


\subsection{Properties Satisfied by SSS Semantics}
We now present various useful properties satisfied by the semantics of AND in Equation~\eqref{add-sub-approx}.


\begin{theorem} [Properties of SSS semantics]
\label{thm:prop}
The STL semantics generated by operator $\mathcal{A}$ in Equation~\eqref{add-sub-approx} is \\
1)~recursive, i.e., the qualitative/quantitative satisfaction of a specification can be derived from the qualitative/quantitative satisfaction of its sub-specifications, \\
2)~satisfies min-max boundedness, i.e. the following condition holds: 
\begin{align*}
\min\,(\rho_1,\ldots,\rho_n)\leq \mathcal{A}(\rho_1,\ldots,\rho_n)\leq \max\,(\rho_1,\ldots,\rho_n),
\end{align*}
3)~continuous and differentiable, and \\ 
4)~satisfies the shadow-lifting property, i.e. for any $\rho \neq 0$,
\begin{align*}
\frac{\partial \mathcal{A}(\rho_1,\ldots,\rho_i,\ldots,\rho_n)}{\partial \rho_i} \mid_{\rho_1,\ldots,\rho_n=\rho} ~>0 , \forall i=1,\ldots,n.
\end{align*}
\end{theorem}

\shortversion{
\begin{proof}
Please see the full version~\cite{fullpaper}.
\end{proof}}
\longversion{
\begin{proof}
\label{pf-prop}
1) As we know, the corresponding max function for the $\min$ function defined in Equation~\eqref{mindef} is given by
\begin{equation}
 \max\,(x_1+x_2) = \frac{(x_1+x_2)+|x_1-x_2|}{2}   
\end{equation}

Consequently, the \texttt{SSS} semantics for OR operator would be defined as:
\begin{equation}
\begin{split}
 \mathcal{O}(\rho_1,\ldots,\rho_n) =  {} & \frac{\sum_{i=1}^{n} \rho_i+\{\delta^{max} \cdot \mathtt{erf}(\mu \cdot \delta^{max})\}}{n}
\end{split}
\end{equation}

De Morgan's law expresses the relation between AND and OR operator as follows
\begin{equation}
    \mathcal{O}(\rho_1,\ldots,\rho_n)=\mathcal{N}(\mathcal{A}(\mathcal{N}(\rho_1),\ldots,\mathcal{N}(\rho_n)))
\end{equation}
where $\mathcal{N}$ refers to negation.

So,
   $\mathcal{N}(\mathcal{A}(\mathcal{N}(\rho_1),\ldots,\mathcal{N}(\rho_n)))$ 
  \begin{align*} 
   {} &= \neg(\mathcal{A}(\neg \rho_1,\ldots,\neg \rho_n))\\
    &=\neg\{\frac{-\sum_{i=1}^{n} \rho_i-\{\delta^{max}\cdot \mathtt{erf}(\mu \cdot \delta^{max})\}}{n}\} \\ 
    &= \frac{\sum_{i=1}^{n} \rho_i+\{\delta^{max} \cdot \mathtt{erf}(\mu \cdot \delta^{max})\}}{n} \\
    &= \mathcal{O}(\rho_1,\ldots,\rho_n) 
  \end{align*}
 
2) Since $\mathtt{erf}(x)\in[0,1]$ for $x\in[0,1]$, we can approximate the expression $x \cdot \mathtt{erf}(\mu \cdot x)$ with minimum value $0$ and maximum value $x$.
Let us consider the first part of the inequality 
\begin{align*}
\underset{i}{\min\,}(\rho_i) & \leq \mathcal{A}(\rho_1,\ldots,\rho_n) \\
    &\leq \frac{\sum_{i=1}^{n} \rho_i-\{\delta^{max} \cdot \mathtt{erf}(\mu \cdot \delta^{max})\}}{n} \\
\end{align*}

The minimum value of $\mathcal{A}(\rho_1,\ldots,\rho_n)$ is achieved when expression $\mathtt{erf}(\mu \cdot \delta^{max})$ is at its maximum value, i.e., $1$.

\begin{align*}
\underset{i}{\min\,}(\rho_i)\leq \frac{\sum_{i=1}^{n} \rho_i-(\underset{i}{\max\,}(\rho_i)-\underset{i}{\min\,}(\rho_i))}{n}.
\end{align*}
\begin{align*}
(n-1)\cdot \underset{i}{\min\,}(\rho_i)\leq \sum_{i=1}^{n} \rho_i-(\underset{i}{\max\,}(\rho_i)).
\end{align*}
The RHS of the above inequality represents the sum of $n-1$ robustness values, excluding the largest robustness. This is definitely greater than or equal to $n-1$ times the smallest robustness value. This proves the first part of inequality.

For the second part of the inequality
\begin{align*}
\mathcal{A}(\rho_1,\ldots,\rho_n) \leq \underset{i}{max}(\rho_i)
\end{align*}
The maximum value of $\mathcal{A}(\rho_1,\ldots,\rho_n)$ is achieved when expression $\mathtt{erf}(\mu\cdot \delta^{max})$ is at its minimum value, i.e., $0$.

\begin{align*}
\frac{\sum_{i=1}^{n} \rho_i}{n} \leq \underset{i}{max}(\rho_i)
\end{align*}
\begin{align*}
\sum_{i=1}^{n} \rho_i \leq n \cdot \underset{i}{max}(\rho_i)
\end{align*}
The RHS above is $n$ times the largest robustness value which is always greater than the LHS. 

3) The summation function $\Sigma$ and the error function $\mathtt{erf}()$ are both well known continuous and differentiable functions. We can easily verify that the expression $x\cdot \mathtt{erf}(\mu \cdot x)$ is also differentiable. As we know, the difference between two differentiable expressions is also differentiable. Hence, the operator $\mathcal{A}$ is continuous and differentiable. 

4) Consider the partial derivative of $\mathcal{A}(\rho_1,\ldots,\rho_n)$ w.r.t $\rho_i$
\begin{align*}
    \frac{\partial \mathcal{A}}{\partial \rho_i} &= \underset{\epsilon\rightarrow 0}{lim} \frac{\mathcal{A}(\rho+\epsilon,\rho,\ldots,\rho)-\mathcal{A}(\rho,\ldots,\rho)}{\epsilon} \\
    &= \underset{\epsilon\rightarrow 0}{lim} \frac{\frac{(\rho \cdot n+\epsilon -(\rho+\epsilon-\rho)\cdot \mathtt{erf}(\mu \cdot (\rho+\epsilon-\rho))}{n}-\rho}{\epsilon} \\
    &= \underset{\epsilon\rightarrow 0}{lim} \frac{\frac{\epsilon}{n} \{1-\mathtt{erf}(\mu \cdot \epsilon)\}}{\epsilon} \\
    &= \underset{\epsilon\rightarrow 0}{lim} \frac{ \{1-\mathtt{erf}(\mu \cdot \epsilon)\}}{n} \\
    &= \frac{1}{n} 
  \end{align*}
  
  Since $n$ is positive, so $ \frac{\partial \mathcal{A}}{\partial \rho_i}>0$. ~~~~~~~~~~~~~~~~~~~~~~~~~~~~~~~~~~~~~~~~\qedsymbol
\end{proof}
}

It has been proved in~\cite{varnai20} that the operator AND cannot be sound, idempotent, and smooth simultaneously. Consequently, the authors in~\cite{varnai20} came up with a metric that was sound and idempotent but not smooth. We instead propose a metric that is idempotent and smooth but not sound.
This is because soundness as a property is not relevant when we consider the robustness metric as the reward for enforcing desirable behavior.
For, instance, if the specification is too optimistic, it might not be satisfiable. Yet, by using this specification, we may converge to a desirable behavior. We can achieve the same behavior with (say) two different specifications - one having stricter constraints and the other with relaxed constraints. With learning, we will eventually get the desirable behavior in both cases - former with negative robustness (not sound) and latter with positive robustness (sound). 
Instead, smoothness plays a crucial role in gradients-based controller synthesis methods~\cite{lse}.
Nevertheless, our semantics satisfies the aggregate notion of soundness (Theorem~\ref{thm-sound}), and we provide the bounds on the robustness values generated by the operator in all cases.

Depending on the robustness values being positive or negative, we can split it into three cases - all positive, all negative, and lastly, some positive and some negative.
In the theorem below, we prove that if all individual robustnesses are positive (negative), then the \texttt{SSS} semantics will certainly give us a positive (negative) robustness (satisfies soundness). Now, consider the third case wherein some robustnesses are positive and some are negative. Obviously, in this case, the largest robustness value is $\rho^{max}>0$, and the smallest robustness value is $\rho^{min}<0$. Depending upon the magnitude of the positive and negative robustness values in the set $\rho_1,\ldots,\rho_n$, the sum of the robustness values can be either positive or negative. We provide a bound on the robustness for both these cases as well (here, traditional soundness requires negative robustness for $\mathcal{A}$ in both cases). 
The traditional soundness is based on the \textit{min} function, i.e., for a given set of values, if the minimum is negative, we require robustness of $\mathcal{A}$ to be negative. On the other hand, here we define the notion of aggregate soundness, which is based on all values in the set instead of point-based estimates (i.e., via min function). 
In the below theorem, part 1, 2, and 4 are also applicable in the case of traditional soundness as well while part 3 is unique to aggregate soundness.
In Part 3, traditional soundness would have required robustness of $\mathcal{A}$ to always remain negative since $\rho^{min}$ is negative. However, aggregate soundness takes into account the fact that the overall sum is positive, and hence robustness of $\mathcal{A}$ can be positive if the magnitude of positive robustness values outweigh the negative ones.

\begin{theorem} [Aggregate Soundness]
\label{thm-sound}
Consider the set of robustness values \{$\rho_1,\ldots,\rho_n$\}.
The operator $\mathcal{A}$ in Equation~\eqref{add-sub-approx} is sound in aggregate sense i.e., it satisfies the following conditions:
\begin{align*}
1)~\mathcal{A}(\rho_1,\ldots,\rho_n)>0,  \hskip60pt \text{if} ~\forall i\in[1,n] ~~\rho_i>0\\
2)~\mathcal{A}(\rho_1,\ldots,\rho_n)<0,  \hskip60pt \text{if} ~\forall i\in[1,n] ~~\rho_i<0\\ 
 3)~\mathcal{A}(\rho_1,\ldots,\rho_n)>-\frac{1}{n}[|\rho^{max}|+|\rho^{min}|], \hskip6pt \text{if}~\sum_{i=1}^{n} \rho_i>0,\\ \rho^{max}>0, ~~\rho^{min}<0\\
  4)~\mathcal{A}(\rho_1,\ldots,\rho_n)<0, 
  ~~~~~~~~~~~~~~~~~\text{if}~\sum_{i=1}^{n} \rho_i<0, \rho^{max}>0,\\ \rho^{min}<0.
  \end{align*}
  where $\rho^{max}=\underset{i}{\max\,}\rho_i$ and $\rho^{min}=\underset{i}{\min\,}\rho_i ~\forall i\in[1,n]$.
\end{theorem}


\shortversion{
\begin{proof}
Please see the full version~\cite{fullpaper}.
\end{proof}}
\longversion{
\begin{proof}
\label{pf-sound}
\textit{For Part 1 and 2:} Note that input to the $\mathtt{erf}$ function in Equation~\eqref{add-sub-approx} is always positive (since $\delta^{max}>0$). This results in the $\mathtt{erf}$ value always $\in [0,1]$.
We consider the two extremes of Equation~\eqref{add-sub-approx} with $\mathtt{erf}=0$ and $\mathtt{erf}=1$.
The value of $\mathcal{A}$ lies within the range
\begin{equation}
\label{erf01}
\begin{split}
 \mathcal{A}(\rho_1,\rho_2,\ldots,\rho_n) \in {} &  [ 
  \frac{\sum_{i=1}^{n} \rho_i-(\delta^{max})}{n}, \frac{\sum_{i=1}^{n} \rho_i}{n}]
\end{split}
\end{equation}

The term $\frac{\sum_{i=1}^{n} \rho_i}{n}$ is the mean robustness and the term $\frac{\sum_{i=1}^{n} \rho_i-(\delta^{max})}{n}$ is $\frac{1}{n}$-th of the sum of $n-1$ largest robustness plus the minimum robustness.  
These two extreme values will always be positive (negative) if all the robustness values are positive (negative). 
Hence, if all $\rho_i>0$ or all $\rho_i<0$, then the range of robustness of operator $\mathcal{A}$ in Equation~\eqref{erf01} will always contain positive or negative robustness values, respectively.

\textit{For Part 3:}
Let us define two variables $a,b>0$ such that $a=\rho^{max}$ and $b=-\rho^{min}$.
Consider the case when $\sum_{i=1}^{n} \rho_i>0$. This implies, 
\begin{align*}
\rho^{max}+\sum_{i=1}^{n-2}\rho_i+\rho^{min}>0    
\end{align*}

\begin{align*}
a-b+\sum_{i=1}^{n-2}\rho_i>0
\end{align*}

\begin{equation}
\label{bnd}
\sum_{i=1}^{n-2}\rho_i-2\cdot b>-b-a
\end{equation}


As we know, $Range(\mathtt{erf})=[0,1]$ for a positive domain and the operator $\mathcal{A}$ has $\mathtt{erf}$ with input $\mu\cdot\delta^{max}$ which is always positive (since $\mu>0$ and $\rho^{max}-\rho^{min}>0$). So, the minimum value of $\mathcal{A}$ is achieved for $\mathtt{erf(.)}=1$.
Hence,
\begin{align*}
    \mathcal{A}(\rho_1,\ldots,\rho_n) &= \frac{\sum_{i=1}^{n} \rho_i-\{\delta^{max}\cdot \mathtt{erf}(\mu \cdot \delta^{max})\}}{n}  \\
    &\geq \frac{\sum_{i=1}^{n} \rho_i-\{\rho^{max}-\rho^{min}\}}{n}\\
    & \geq \frac{\sum_{i=1}^{n-2} \rho_i+a-b-\{a+b\}}{n}  \\
    & \geq \frac{\sum_{i=1}^{n-2} \rho_i-2\cdot b}{n}  \\
    & > \frac{-[b+a]}{n} \hskip20pt //Using~\eqref{bnd} \\
     & > -\frac{[|\rho^{max}|+|\rho^{min}|]}{n} \\
  \end{align*}

\textit{For Part 4:}
As seen in part 3, $Range(\mathtt{erf})=[0,1]$ for a positive domain. So, maximum value for $\mathcal{A}$ is attained at $\mathtt{erf}(.)=0$. Therefore,
\begin{align*}
    \mathcal{A}(\rho_1,\ldots,\rho_n) &= \frac{\sum_{i=1}^{n} \rho_i-\{\delta^{max}\cdot \mathtt{erf}(\mu \cdot \delta^{max})\}}{n}  \\
    &\leq \frac{\sum_{i=1}^{n} \rho_i}{n}\\
     &< 0\\
\end{align*}
Since for this case $\sum_{i=1}^{n} \rho_i <0$, so the above expression is always negative.
\end{proof}
}

\begin{table}[t]
\begin{center}
\scalebox{0.85}{
\begin{tabular} {lrr}
\toprule
   Environment & \#Observations & \#Actions \\
  \midrule
  HalfCheetah  \textsf{(HC)} & 17 & 6\\
  Hopper \textsf{(Hop)} & 11 & 3 \\
  Ant & 27 & 8\\
  Walker \textsf{(Wkr)} & 17 & 6 \\
  Swimmer \textsf{(Swm)} & 8 & 2\\
  Humanoid \textsf{(Hum)} & 376 & 17 \\
 \bottomrule
\end{tabular}}
\caption{Benchmarks}
\label{table-bench}
\end{center}
\end{table}

\begin{table*}
\begin{center}
\scalebox{0.85}{
\begin{tabular} {lll}
\toprule
  Model & STL Specification & Variables\\
  \midrule
 \textsf{HC} &  $ \lozenge_{[0,10]} (v[t]>0.1)$ & $v$ : velocity \\
  \addlinespace
  \textsf{Hop} 
 & $\lozenge_{[0,15]}(v[t]>0.5) \land \square_{[0,20]} ((z[t]>0.7) \land  (|a[t]|<1))$ & $v$: velocity, $z$: height, $a$: angle \\
 \addlinespace
  \textsf{Ant} 
& $\lozenge_{[0,5]}(vx[t]>0.2) \land \square_{[0,10]} ((z[t]<1) \land (z[t]>0.2))$ & $vx$: $x$-velocity, $z$: height\\
 \addlinespace
\textsf{Wkr}
    & $\lozenge_{[0,15]}(vx[t]>0.5) \land \square_{[0,20]} ((z[t]>0.8) \land (z[t]<2) \land (|a[t]|<1))$ &  $vx$: $x$-velocity, $z$ : height, $a$: angle \\
      \addlinespace
    \textsf{Swm} &  $\lozenge_{[0,15]} ((vx[t]>0.1) \land \square_{[0,20]} ((|a_1[t]|<1) \land (|a_2[t]|<1) \land (|a_3[t]|<1))$ & \multirow{2}{*}{$vx$: $x$-velocity, $\langle a_1,a_2,a_3\rangle$: angles of }\\
       \addlinespace
    & & first tip, first rotor and second rotor \\
   \addlinespace
    \textsf{Hum} &  $\lozenge_{[0,5]}(vx[t]>0.5) \land \square_{[0,5]}((z[t]>1) \land (z[t]<2))$ & $vx$: $x$-velocity, $z$: height \\
 \addlinespace
 \bottomrule
\end{tabular}}
\caption{STL specifications}
\label{table-spec}
\end{center}
\end{table*}

\shortversion{
\begin{table}[t]
\begin{center}
\medskip
\scalebox{0.76}{
\begin{tabular} {llrrrrr}
\toprule
   Env & Sem & CC & DC &  MoS &  SAT & DR\\
  \midrule
  \multirow{6}{*}{\textsf{HC}} & Classical &  $16127.17$  & $307.66$ &  $7.30$ &  - & $5755.00$\\
 & \texttt{AGM} & $16599.73$  & $406.29$ & $10.03$ &  - & $7716.32$ \\
 & \texttt{LSE} & $15311.38$  & $111.86$ & $4.25$ &  - & $1850.36$ \\
 & \texttt{Softmax} & $14154.17$  & $471.29$ & $10.76$ &  - & $9047.92$\\
 & \texttt{SSS} & $14647.78$ & $541.90$ & $12.04$ &  - & $10453.96$\\
 \midrule
 \multirow{6}{*}{\textsf{Hop}} & Classical &  $1698.20$ & $12.08$ & $0.68$ &  $99/100$ & $2505.08$ \\
 & \texttt{AGM} & $928.52$ & $12.72$ & $0.58$ &  $0/100$ & $2196.76$\\
 & \texttt{LSE} & $42.81$  & $0.02$ & $-0.26$ &  $0/100$ & $31.91$\\
 & \texttt{SoftMax} & $1571.84$ & $13.99$ & $0.69$ &  $100/100$ & $2745.39$\\
 & \texttt{SSS} & $1231.04$ & $17.25$ & $0.69$ &  $100/100$ & $3153.21$\\
 \midrule
   \multirow{6}{*}{\textsf{Ant}} & Cls & $12055.73$  & $34.97$ & $0.32$ & $99/100$  & 13.95 \\
  & \texttt{AGM} & $9089.84$  & $40.04$ & $0.31$ & $100/100$ & 366.71 \\
  & \texttt{LSE} & $220.42$  & $0.32$ & $0.21$ & $100/100$ & -11.70 \\
  & \texttt{SMax} & $10938.52$ & $38.61$ & $0.32$  & $100/100$ & 150.44 \\
  & \texttt{SSS} & $12678.64$ & $85.03$ & $0.30$  & $100/100$  & 967.26\\
  \midrule
  \multirow{6}{*}{\textsf{Wkr}}  
  & Classical & $5614.93$ & $8.56$ & $0.71$ &  $99/100$  & $2056.56$ \\
  & \texttt{AGM} & $76.63$ & $-0.07$ & $-0.37$ &  $0/100$ & $-1.54$\\ 
  & \texttt{LSE} & $368.15$ & $-0.08$  & $-0.14$ &  $0/100$ & $44.25$\\
  & \texttt{SoftMax} & $10359.56$ & $-4.53$ & $0.33$ &  $94/100$ & $413.97$\\ 
  & \texttt{SSS} & $10743.50$ & $18.71$  & $0.59$ &  $100/100$ & $3333.06$\\
  \midrule
  \multirow{6}{*}{\textsf{Swm}}  
  & Cls & $626.38$ & $-1.59$ & $0.37$ & $100/100$ & -40.05 \\
  & \texttt{AGM} & $87.48$ & $0.78$ & $0.00$ &  $100/100$ & 19.46\\ 
  & \texttt{LSE} & $56.31$ & $0.25$  & $-0.06$ &  $100/100$ & 6.35\\
  & \texttt{SMax} & $573.94$ & $0.05$ & $0.07$ & $3/100$ & 0.98\\ 
  & \texttt{SSS} & $1256.99$ & $7.18$ & $0.37$ & $99/100$ & 179.09 \\
\midrule
  \multirow{6}{*}{\textsf{Hum}}  
  & Classical & $2072.98$ & $6.66$ & $0.28$ &  $99/100$ & $5396.01$ \\
  & \texttt{AGM} & $3143.56$ & $18.43$ & $0.24$ &  $100/100$ & $6352.30$ \\ 
  & \texttt{LSE} & $1531.31$ & $2.25$  & $0.09$ &  $100/100$ & $5060.10$ \\
  & \texttt{SoftMax} & $2313.87$ & $7.94$ & $0.28$ &  $100/100$ & $5505.03$ \\ 
  & \texttt{SSS} & $2901.91$ & $26.79$ & $0.27$ &  $100/100$ & $7054.92$ \\
 \bottomrule
\end{tabular}}
\caption{Comparison of different semantics under different environments}
\label{table-all-short}
\end{center}
\end{table}
}

\longversion{
\begin{table*}[ht!]
\begin{center}
\scalebox{0.8}{
\begin{tabular} {llrrrrr}
\toprule
   Environment & Semantics & Control Cost & Distance Covered &  $MoS$ &  $SAT$ & $Default Reward$\\
  \midrule
  \multirow{6}{*}{\textsf{HalfCheetah}} & Classical &  $16127.17\pm195.61$  & $307.66\pm4.17$ &  $7.30\pm0.07$ &  - & $5755.00\pm82.57$\\
 & \texttt{AGM} & $16599.73\pm181.25$  & $406.29\pm5.44$ & $10.03\pm0.10$ &  - & $7716.32\pm108.02$ \\
 & \texttt{LSE} & $15311.38\pm100.39$  & $111.86\pm1.26$ & $4.25\pm0.02$ &  - & $1850.36\pm25.09$ \\
 & \texttt{Softmax} & $14154.17\pm198.75$  & $471.29\pm7.10$ & $10.76\pm0.13$ &  - & $9047.92\pm140.68$\\
 & \texttt{SSS} & $14647.78\pm984.72$ & $541.90\pm38.50$ & $12.04\pm0.78$ &  - & $10453.96\pm779.36$\\
 \midrule
 \multirow{6}{*}{\textsf{Hopper}} & Classical &  $1698.20\pm53.72$ & $12.08\pm0.25$ & $0.68\pm0.00$ &  $99/100$ & $2505.08\pm40.64$ \\
 & \texttt{AGM} & $928.52\pm122.87$ & $12.72\pm1.76$ & $0.58\pm0.01$ &  $0/100$ & $2196.76\pm311.55$\\
 & \texttt{LSE} & $42.81\pm5.72$  & $0.02\pm0.02$ & $-0.26\pm0.03$ &  $0/100$ & $31.91\pm4.27$\\
 & \texttt{SoftMax} & $1571.84\pm46.85$ & $13.99\pm0.21$ & $0.69\pm0.00$ &  $100/100$ & $2745.39\pm27.05$\\
 & \texttt{SSS} & $1231.04\pm19.29$ & $17.25\pm0.07$ & $0.69\pm0.00$ &  $100/100$ & $3153.21\pm8.84$\\
 \midrule
  \multirow{6}{*}{\textsf{Ant}} & Classical & $12055.73\pm355.69$  & $34.97\pm1.11$ & $0.32\pm0.00$ &  $99/100$  & $13.95\pm37.32$ \\
  & \texttt{AGM} & $9089.84\pm182.02$  & $40.04\pm0.78$ & $0.31\pm0.00$ &  $100/100$ & $366.71\pm14.32$ \\
  & \texttt{LSE} & $220.42\pm74.78$  & $0.32\pm0.32$ & $0.21\pm0.06$ &  $100/100$ & $-11.70\pm7.54$ \\
  & \texttt{SoftMax} & $10938.52\pm130.68$ & $38.61\pm0.88$ & $0.32\pm0.00$  &  $100/100$ &  $150.44\pm14.73$\\
  & \texttt{SSS} & $12678.64\pm225.12$ & $85.03\pm1.57$ & $0.30\pm0.00$  &  $100/100$ & $967.26\pm36.99$\\
  \midrule
  \multirow{6}{*}{\textsf{Walker}}  
  & Classical & $5614.93\pm456.56$ & $8.56\pm0.81$ & $0.71\pm0.09$ &  $99/100$  & $2056.56\pm188.54$ \\
  & \texttt{AGM} & $76.63\pm1.25$ & $-0.07\pm0.00$ & $-0.37\pm0.02$ &  $0/100$ & $-1.54\pm0.07$\\ 
  & \texttt{LSE} & $368.15\pm59.34$ & $-0.08\pm0.01$  & $-0.14\pm0.07$ &  $0/100$ & $44.25\pm7.59$\\
  & \texttt{SoftMax} & $10359.56\pm700.12$ & $-4.53\pm0.87$ & $0.33\pm0.03$ &  $94/100$ & $413.97\pm114.76$\\ 
  & \texttt{SSS} & $10743.50\pm189.36$ & $18.71\pm0.11$  & $0.59\pm0.00$ &  $100/100$ & $3333.06\pm14.39$\\
  \midrule
  \multirow{6}{*}{\textsf{Swimmer}}  
  & Classical & $626.38\pm38.65$ & $-1.59\pm0.21$ & $0.37\pm0.02$ &  $100/100$ & $-40.05\pm5.24$\\
  & \texttt{AGM} & $87.48\pm51.06$ & $0.78\pm0.36$ & $0.00\pm0.02$ &  $100/100$ & $19.46\pm9.06$\\ 
  & \texttt{LSE} & $56.31\pm76.83$ & $0.25\pm0.67$  & $-0.06\pm0.02$ &  $100/100$ & $6.35\pm16.89$\\
  & \texttt{SoftMax} & $573.94\pm78.30$ & $0.05\pm0.19$ & $0.07\pm0.10$ &  $3/100$ & $0.98\pm4.90$\\ 
  & \texttt{SSS} & $1256.99\pm16.22$ & $7.18\pm0.12$ & $0.37\pm0.00$ &  $99/100$  & $179.09\pm2.89$\\
\midrule
  \multirow{6}{*}{\textsf{Humanoid}}  
  & Classical & $2072.98\pm37.73$ & $6.66\pm0.39$ & $0.28\pm0.00$ &  $99/100$ & $5396.01\pm119.59$ \\
  & \texttt{AGM} & $3143.56\pm35.54$ & $18.43\pm0.15$ & $0.24\pm0.00$ &  $100/100$ & $6352.30\pm12.85$ \\ 
  & \texttt{LSE} & $1531.31\pm114.41$ & $2.25\pm1.03$  & $0.09\pm0.07$ &  $100/100$ & $5060.10\pm81.72$ \\
  & \texttt{SoftMax} & $2313.87\pm37.79$ & $7.94\pm0.09$ & $0.28\pm0.00$ &  $100/100$ & $5505.03\pm7.43$ \\ 
  & \texttt{SSS} & $2901.91\pm31.20$ & $26.79\pm0.55$ & $0.27\pm0.00$ &  $100/100$ & $7054.92\pm45.94$ \\
 \bottomrule
\end{tabular}}
\caption{Comparison of different semantics under different environments}
\label{table-all-short}
\end{center}
\end{table*}
}

\section{Experiments}

This section present our experiments to evaluate the Deep-RL based controller synthesis methodology for STL specifications using our proposed aggregation based semantics. 
All experiments are carried out on an Ubuntu20.04 machine with i7-4770 3.40\,GHz$\times$8 CPU and 32\,GB RAM.
The source code of our implementation and the videos of the simulation results are at \url{https://github.com/iitkcpslab/rlstl}.

\subsection{Experimental Setup}

\subsubsection{Implementation.}
The controllers used in our experiments are synthesized in a Deep-RL 
setting with continuous state space and continuous action space. To simulate the environment, we use the $\texttt{gym}$ simulator~\cite{gym}. 
The learning is performed using the state-of-the-art Actor-Critic algorithm SAC~\cite{haarnoja18}. We modify the SAC algorithm to use the STL robustness computed online as the reward.
We implement different STL semantics on top of the online monitoring tool RTAMT~\cite{rtamt}. We have also developed a Python program for controller evaluation in the \texttt{gym} environment.




We have performed our experiments on six continuous control benchmarks, namely HalfCheetah~\cite{hc_ref} (\textsf{HC}), Hopper~\cite{hopper} (\textsf{Hop}), \textsf{Ant}~\cite{ant}, Walker~\cite{hopper} (\textsf{Wkr}), Swimmer~\cite{swimmer} (\textsf{Swm}), and Humanoid~\cite{hopper} (\textsf{Hum}).
The number of observable states and actions for the benchmarks are shown in Table~\ref{table-bench}.
In Table~\ref{table-spec}, we list the specifications used for training the controllers.
All the units are in SI.
Each specification captures the goal defined for the system by the \texttt{gym} environment. 
Each specification has a liveness component and a safety component (other than \textsf{HC}). 
For instance, for the Hopper example, the goal is to ``Make a two-dimensional one-legged robot hop forward as fast as possible without falling''. This is captured by the specification
\begin{small}
\[
  \phi =  
\underbrace{\black{\lozenge_{[0,15]} (v[t]>0.5)}}_{\text{liveliness}} \land
  \underbrace{\black{\square_{[0,20]}((z[t]>0.7) \land (abs(a)<1)}}_{\text{safety}}).
\]
\end{small}
The meaning of this specification should be understood in terms of the robustness that would be generated w.r.t. this specification which gives us the reward for moving towards the goal. 
Intuitively, the liveness component of the specification implies that from any time step, the hopper should try to achieve velocity of at least 0.5 units within the next $15$ time steps. For achieving velocity less than, equal to, and greater than 0.5 units, the reward (robustness) it will get will be negative, zero, and positive, respectively.
The safety component ensures that the hopper never falls during the forward movement. 
All the safety requirements used in our experiments are exactly the same as that in the corresponding $\texttt{gym}$ environment.



\subsubsection{Hyper-parameters.}
There are different hyper-parameters used in various semantics of STL. We have used the hyperparameter values ($\beta=1$, $\nu=3$) used in the work~\cite{lse,varnai20}. 
For our \texttt{SSS} semantics, we carry out initial experiments on the Hopper benchmark to understand the effect of the hyperparameters $\mu$ and $\eta$ on the performance of the controller. Based on the insights obtained from experiments, we choose $\mu=0.3$ and $\eta=300$ for all our experiments.
The benchmark configuration hyper-parameters, such as learning rate, total time-steps for training, etc., are taken from Stable-baselines3~\cite{stable-baselines3}.

\subsection{Experimental Results}

Table~\ref{table-all-short} presents the performance of the controllers synthesized using different semantics for various benchmarks.
We tested the synthesized controller on 100 distinct seed inputs for better coverage of the behavior of these controllers.
\shortversion{
In Table~\ref{table-all-short}, we show the mean values (100 seeds) for all the metrics. The complete results (along with standard deviation) can be found in~\cite{fullpaper}. }
\longversion{
In Table~\ref{table-all-short}, we show the mean and standard deviation (100 seeds) for all the metrics.}
The results show that our proposed STL semantics $\texttt{SSS}$ outperforms all the other semantics on the metrics introduced in Section~\ref{sec-evaluation-metrics}.

Some key observations about the experimental results shown in Table~\ref{table-all-short} are in order.
(i) \texttt{SSS} is the only semantics that could synthesize a useful controller for all the benchmarks. In the case of Swimmer, a useful controller (enables the robot to move gracefully in the forward direction without violating the safety property) could be synthesized only via \texttt{SSS} robustness semantics. 
(ii) The controller synthesized by \texttt{SSS} achieves the maximum distance covered (DC) among all the controllers. It is also the best for acquiring \texttt{gym}'s default reward (DR).
(iii) In many cases, the controller synthesized using \texttt{SSS} robustness was not the one with minimum control cost (CC). This is reasonable because often low control cost (CC) means that the controller is not generating controls for the movement. For instance, in the case of Swimmer, for \texttt{LSE} and \texttt{AGM}, we have low control cost, but that is because the Swimmer is not moving at all. 
On the other hand, despite the high control cost for \texttt{classical} and \texttt{Softmax}, the Swimmer is not able to move. 
This shows that the controllers do not produce the correct control inputs.
(iv) In almost all cases, the controller corresponding to \texttt{SSS} semantics has the highest MoS except for Ant and Humanoid. In both these cases, slightly less than the highest MoS is achieved. This is because it has to perform a movement with a lower safety margin for faster movement. 
This is reasonable as ideally the safety requirement has to be satisfied by any satisfaction margin.
(v) In almost all cases, we get the best controller in terms of SAT with $100/100$ satisfaction of the safety specification. In case of Swimmer, due to stricter safety requirements, a useful controller was synthesized only via \texttt{SSS} semantics. However, this was achieved at the cost of 99/100 safety satisfaction. It is worth noting that the safety constraints are easy to satisfy if the Swimmer does not move forward significantly. 
This is the reason why the \texttt{Classical}, \texttt{AGM}, and \texttt{LSE} semantics can achieve 100/100 satisfaction of the safety property. The \texttt{Softmax} semantics performs the worse, failing to satisfy both the liveness and the safety requirements, owing to the U-shape motion of the Swimmer with the resulting controller, leading to poor DC and many safety violations.  
\section{Conclusion}
We propose a new aggregation-based smooth STL semantics for synthesizing feedback controllers via Reinforcement Learning. Our STL semantics is not sound, but has other desirable properties required for efficient RL based controller synthesis. 
We provide a comparative analysis of the proposed semantics with all the state-of-the-art STL semantics. 
Experimental results on several complex benchmarks establish that our proposed semantics outperforms all the aggregate-based STL semantics proposed in the literature.  
Going forward, we plan to explore the application of our semantics in other complex domains like self-driving cars.

\section* {Acknowledgements}
The authors would like to thank anonymous reviewers for their valuable feedback. The authors also thank Dejan Nickovic for his
support in working with the RTAMT Toolbox.


\begin{thebibliography}{36}
\providecommand{\natexlab}[1]{#1}

\bibitem[{Alpern and Schneider(1987)}]{Alpern87}
Alpern, B.; and Schneider, F.~B. 1987.
\newblock Recognizing Safety and Liveness.
\newblock \emph{Distrib. Comput.}, 2(3): 117–126.

\bibitem[{Bagul and Chesneau(2020)}]{Bagul20}
Bagul, Y.~J.; and Chesneau, C. 2020.
\newblock Sigmoid functions for the smooth approximation to the absolute value
  function.
\newblock \emph{Moroccan Journal of Pure and Applied Analysis}, 7: 12 -- 19.

\bibitem[{Brockman et~al.(2016)Brockman, Cheung, Pettersson, Schneider,
  Schulman, Tang, and Zaremba}]{gym}
Brockman, G.; Cheung, V.; Pettersson, L.; Schneider, J.; Schulman, J.; Tang,
  J.; and Zaremba, W. 2016.
\newblock OpenAI Gym.
\newblock \emph{CoRR}, abs/1606.01540.

\bibitem[{Coulom(2002)}]{swimmer}
Coulom, R. 2002.
\newblock \emph{Reinforcement Learning Using Neural Networks, with Applications
  to Motor Control}.
\newblock Ph.D. thesis, Institut National Polytechnique de Grenoble.

\bibitem[{Deisenroth, Neumann, and Peters(2013)}]{Deisenroth13}
Deisenroth, M.~P.; Neumann, G.; and Peters, J. 2013.
\newblock A Survey on Policy Search for Robotics.
\newblock \emph{Found. Trends Robotics}, 2: 1--142.

\bibitem[{Donz{\'e}, Ferr{\`e}re, and Maler(2013)}]{Donxe13}
Donz{\'e}, A.; Ferr{\`e}re, T.; and Maler, O. 2013.
\newblock Efficient Robust Monitoring for {STL}.
\newblock In \emph{CAV}, 264--279.

\bibitem[{Donz{\'e} and Maler(2010)}]{donze10a}
Donz{\'e}, A.; and Maler, O. 2010.
\newblock Robust Satisfaction of Temporal Logic over Real-Valued Signals.
\newblock In \emph{Formal Modeling and Analysis of Timed Systems}, 92--106.

\bibitem[{Erez, Tassa, and Todorov(2011)}]{hopper}
Erez, T.; Tassa, Y.; and Todorov, E. 2011.
\newblock Infinite-Horizon Model Predictive Control for Periodic Tasks with
  Contacts.
\newblock In \emph{Robotics: Science and Systems}.

\bibitem[{Fulton and Platzer(2018)}]{Fulton18}
Fulton, N.; and Platzer, A. 2018.
\newblock Safe Reinforcement Learning via Formal Methods: Toward Safe Control
  Through Proof and Learning.
\newblock \emph{AAAI}, 32(1).

\bibitem[{Haarnoja et~al.(2018)Haarnoja, Zhou, Abbeel, and Levine}]{haarnoja18}
Haarnoja, T.; Zhou, A.; Abbeel, P.; and Levine, S. 2018.
\newblock Soft Actor-Critic: Off-Policy Maximum Entropy Deep Reinforcement
  Learning with a Stochastic Actor.
\newblock In \emph{ICML}, 1861--1870.

\bibitem[{Hafner and Riedmiller(2011)}]{Hafner2011}
Hafner, R.; and Riedmiller, M.~A. 2011.
\newblock Reinforcement learning in feedback control.
\newblock \emph{Machine Learning}, 84: 137--169.

\bibitem[{Jaksic et~al.(2018)Jaksic, Bartocci, Grosu, Nguyen, and
  Ni{\v{c}}kovi{\'{c}}}]{jaksic18}
Jaksic, S.; Bartocci, E.; Grosu, R.; Nguyen, T.; and Ni{\v{c}}kovi{\'{c}}, D.
  2018.
\newblock Quantitative Monitoring of {STL} with Edit Distance.
\newblock \emph{Form. Methods Syst. Des.}, 53(1): 83–112.

\bibitem[{Kindler(1994)}]{Kindler94}
Kindler, E. 1994.
\newblock Safety and Liveness Properties: {A} Survey.
\newblock \emph{EATCS-Bulletin}, 53.

\bibitem[{Konda and Tsitsiklis(2003)}]{Konda03}
Konda, V.~R.; and Tsitsiklis, J.~N. 2003.
\newblock OnActor-Critic Algorithms.
\newblock \emph{SIAM Journal on Control and Optimization}, 42(4): 1143--1166.

\bibitem[{Lamport(2000)}]{Lamport00}
Lamport, L. 2000.
\newblock Fairness and Hyperfairness.
\newblock \emph{Distrib. Comput.}, 13(4): 239–245.

\bibitem[{Levine et~al.(2016)Levine, Finn, Darrell, and Abbeel}]{Levine16}
Levine, S.; Finn, C.; Darrell, T.; and Abbeel, P. 2016.
\newblock End-to-End Training of Deep Visuomotor Policies.
\newblock \emph{J. Mach. Learn. Res.}, 17(1): 1334–1373.

\bibitem[{Levine and Koltun(2013)}]{levine13}
Levine, S.; and Koltun, V. 2013.
\newblock Guided Policy Search.
\newblock In \emph{ICML}, volume~28 of \emph{Proceedings of Machine Learning
  Research}, 1--9. Atlanta, Georgia, USA: PMLR.

\bibitem[{Li, Ma, and Belta(2018)}]{lse}
Li, X.; Ma, Y.; and Belta, C. 2018.
\newblock A Policy Search Method For Temporal Logic Specified Reinforcement
  Learning Tasks.
\newblock In \emph{ACC}, 240--245.

\bibitem[{Lillicrap et~al.(2016)Lillicrap, Hunt, Pritzel, Heess, Erez, Tassa,
  Silver, and Wierstra}]{lillicrap19}
Lillicrap, T.~P.; Hunt, J.~J.; Pritzel, A.; Heess, N.; Erez, T.; Tassa, Y.;
  Silver, D.; and Wierstra, D. 2016.
\newblock Continuous control with deep reinforcement learning.
\newblock In \emph{ICLR}.

\bibitem[{Mehdipour, Vasile, and Belta(2019)}]{agm}
Mehdipour, N.; Vasile, C.~I.; and Belta, C. 2019.
\newblock Arithmetic-Geometric Mean Robustness for Control from Signal Temporal
  Logic Specifications.
\newblock In \emph{ACC}, 1690--1695. {IEEE}.

\bibitem[{Mnih et~al.(2015)Mnih, Kavukcuoglu, Silver, Rusu, Veness, Bellemare,
  Graves, Riedmiller, Fidjeland, Ostrovski, Petersen, Beattie, Sadik,
  Antonoglou, King, Kumaran, Wierstra, Legg, and Hassabis}]{Mnih15}
Mnih, V.; Kavukcuoglu, K.; Silver, D.; Rusu, A.~A.; Veness, J.; Bellemare,
  M.~G.; Graves, A.; Riedmiller, M.~A.; Fidjeland, A.; Ostrovski, G.; Petersen,
  S.; Beattie, C.; Sadik, A.; Antonoglou, I.; King, H.; Kumaran, D.; Wierstra,
  D.; Legg, S.; and Hassabis, D. 2015.
\newblock Human-level control through deep reinforcement learning.
\newblock \emph{Nature}, 518: 529--533.

\bibitem[{Nachum et~al.(2018)Nachum, Norouzi, Tucker, and
  Schuurmans}]{nachum18}
Nachum, O.; Norouzi, M.; Tucker, G.; and Schuurmans, D. 2018.
\newblock Smoothed Action Value Functions for Learning {G}aussian Policies.
\newblock In \emph{Proceedings of the 35th International Conference on Machine
  Learning}, volume~80, 3692--3700. PMLR.

\bibitem[{Ni{\v{c}}kovi{\'{c}} and Yamaguchi(2020)}]{rtamt}
Ni{\v{c}}kovi{\'{c}}, D.; and Yamaguchi, T. 2020.
\newblock {RTAMT:} {O}nline Robustness Monitors from {STL}.
\newblock In \emph{Automated Technology for Verification and Analysis},
  564--571.

\bibitem[{Pant, Abbas, and Mangharam(2017)}]{Pant17}
Pant, Y.~V.; Abbas, H.; and Mangharam, R. 2017.
\newblock Smooth operator: Control using the smooth robustness of temporal
  logic.
\newblock In \emph{2017 IEEE Conference on Control Technology and Applications
  (CCTA)}, 1235--1240.

\bibitem[{Raffin et~al.(2021)Raffin, Hill, Gleave, Kanervisto, Ernestus, and
  Dormann}]{stable-baselines3}
Raffin, A.; Hill, A.; Gleave, A.; Kanervisto, A.; Ernestus, M.; and Dormann, N.
  2021.
\newblock Stable-Baselines3: Reliable Reinforcement Learning Implementations.
\newblock \emph{Journal of Machine Learning Research}, 22(268): 1--8.

\bibitem[{Raman et~al.(2015)Raman, Donz\'{e}, Sadigh, Murray, and
  Seshia}]{raman15}
Raman, V.; Donz\'{e}, A.; Sadigh, D.; Murray, R.~M.; and Seshia, S.~A. 2015.
\newblock Reactive Synthesis from Signal Temporal Logic Specifications.
\newblock In \emph{HSCC}, 239–248.

\bibitem[{{Raman} et~al.(2014){Raman}, {Donzé}, {Maasoumy}, {Murray},
  {Sangiovanni-Vincentelli}, and {Seshia}}]{raman14}
{Raman}, V.; {Donzé}, A.; {Maasoumy}, M.; {Murray}, R.~M.;
  {Sangiovanni-Vincentelli}, A.; and {Seshia}, S.~A. 2014.
\newblock Model predictive control with signal temporal logic specifications.
\newblock In \emph{CDC}, 81--87.

\bibitem[{Schulman et~al.(2016)Schulman, Moritz, Levine, Jordan, and
  Abbeel}]{ant}
Schulman, J.; Moritz, P.; Levine, S.; Jordan, M.; and Abbeel, P. 2016.
\newblock High-Dimensional Continuous Control Using Generalized Advantage
  Estimation.
\newblock In \emph{ICLR}.

\bibitem[{Shen et~al.(2020)Shen, Li, Jiang, Wang, and Zhao}]{shen20}
Shen, Q.; Li, Y.; Jiang, H.; Wang, Z.; and Zhao, T. 2020.
\newblock Deep Reinforcement Learning with Robust and Smooth Policy.
\newblock In \emph{Proceedings of the 37th International Conference on Machine
  Learning (ICML)}. JMLR.

\bibitem[{Silver et~al.(2017)Silver, Schrittwieser, Simonyan, Antonoglou,
  Huang, Guez, Hubert, baker, Lai, Bolton, Chen, Lillicrap, Hui, Sifre, van~den
  Driessche, Graepel, and Hassabis}]{Silver17}
Silver, D.; Schrittwieser, J.; Simonyan, K.; Antonoglou, I.; Huang, A.; Guez,
  A.; Hubert, T.; baker, L.; Lai, M.; Bolton, A.; Chen, Y.; Lillicrap, T.~P.;
  Hui, F.; Sifre, L.; van~den Driessche, G.; Graepel, T.; and Hassabis, D.
  2017.
\newblock Mastering the game of Go without human knowledge.
\newblock \emph{Nature}, 550: 354--359.

\bibitem[{Singh and Saha(2021)}]{Singh21}
Singh, N.~K.; and Saha, I. 2021.
\newblock Specification Guided Automated Synthesis of Feedback Controllers.
\newblock \emph{ACM Trans. Embed. Comput. Syst.}, 20(5).

\bibitem[{Singh and Saha(2022)}]{fullpaper}
Singh, N.~K.; and Saha, I. 2022.
\newblock STL-Based Synthesis of Feedback Controllers Using Reinforcement
  Learning.
\newblock \emph{CoRR}, abs/2103.02821.

\bibitem[{Sutton and Barto(2018)}]{sutton}
Sutton, R.~S.; and Barto, A.~G. 2018.
\newblock \emph{Reinforcement Learning: An Introduction}.
\newblock Cambridge, MA, USA: A Bradford Book.
\newblock ISBN 0262039249.

\bibitem[{Tassa, Erez, and Todorov(2012)}]{tassa12}
Tassa, Y.; Erez, T.; and Todorov, E. 2012.
\newblock Synthesis and stabilization of complex behaviors through online
  trajectory optimization.
\newblock In \emph{IROS}.

\bibitem[{V{\'a}rnai and Dimarogonas(2020)}]{varnai20}
V{\'a}rnai, P.; and Dimarogonas, D.~V. 2020.
\newblock On Robustness Metrics for Learning {STL} Tasks.
\newblock In \emph{ACC}, 5394--5399.

\bibitem[{Wawrzy{\'{n}}ski(2009)}]{hc_ref}
Wawrzy{\'{n}}ski, P. 2009.
\newblock A Cat-Like Robot Real-Time Learning to Run.
\newblock In Kolehmainen, M.; Toivanen, P.; and Beliczynski, B., eds.,
  \emph{Adaptive and Natural Computing Algorithms}, 380--390. Berlin,
  Heidelberg: Springer Berlin Heidelberg.
\newblock ISBN 978-3-642-04921-7.

\end{thebibliography}
\end{document}